\pdfoutput=1
\documentclass[11pt]{article}

\usepackage[preprint]{acl}

\usepackage{times}
\usepackage{latexsym}

\usepackage[T1]{fontenc}
\usepackage[utf8]{inputenc}

\usepackage{microtype}

\usepackage{inconsolata}

\usepackage{graphicx}
\usepackage{amsmath}
\usepackage{booktabs}
\usepackage{xcolor}

\title{Think-then-Act: A Dual-Angle Evaluated Retrieval-Augmented Generation}

\author{Yige Shen \\ Xi'an Jiaotong University \\ shenyige@stu.xjtu.edu.cn
        \And  
        Hao Jiang \\ Xi'an Jiaotong University \\ haojiang@stu.xjtu.edu.cn
        \AND
        Hua Qu \\ Xi'an Jiaotong University \\ qh@mail.xjtu.edu.cn
        \And
        Jihong Zhao \\ Xi'an Jiaotong University \\ zhaojihong@mail.xjtu.edu.cn
        }

\begin{document}
\maketitle
\begin{abstract}
Despite their impressive capabilities, large language models (LLMs) often face challenges such as temporal misalignment and generating hallucinatory content. Enhancing LLMs with retrieval mechanisms to fetch relevant information from external sources offers a promising solution.
Inspired by the proverb "Think twice before you act," we propose a dual-angle evaluated retrieval-augmented generation framework \textit{Think-then-Act}. Unlike previous approaches that indiscriminately rewrite queries or perform retrieval regardless of necessity, or generate temporary responses before deciding on additional retrieval, which increases model generation costs, our framework employs a two-phase process: (i) assessing the input query for clarity and completeness to determine if rewriting is necessary; and (ii) evaluating the model's capability to answer the query and deciding if additional retrieval is needed.
Experimental results on five datasets show that the \textit{Think-then-Act} framework significantly improves performance. Our framework demonstrates notable improvements in accuracy and efficiency compared to existing baselines and performs well in both English and non-English contexts. Ablation studies validate the optimal model confidence threshold, highlighting the resource optimization benefits of our approach.
\end{abstract}

\begin{figure*}[t]
  \centering
  \includegraphics[width=0.89\linewidth]{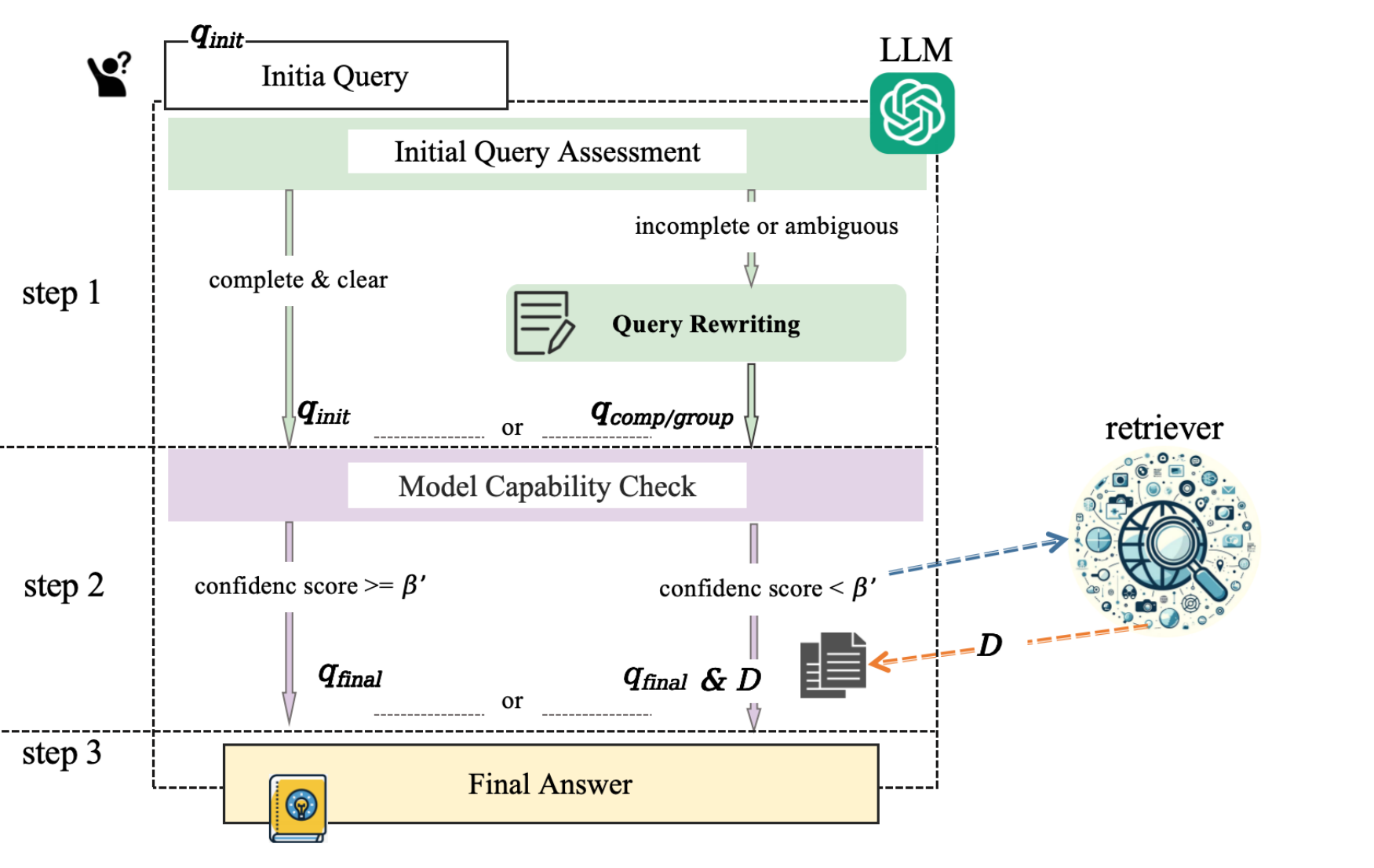}
  \caption {Think-then-Act: (i) assessing the input query for clarity and completeness to determine if rewriting is necessary; (ii) evaluating the model's capability to answer the query and deciding if additional retrieval is needed.}
  \label{fig: frameword}
\end{figure*}

\section{Introduction}
Large language models (LLMs) have become a cornerstone of natural language processing (NLP) systems due to their impressive capabilities in understanding and generating human language \citep{brown2020language,ouyang2022training,openai2023gpt}.
Despite their success, LLMs often suffer from temporal misalignment \cite{rottger-pierrehumbert-2021-temporal-adaptation,luu-etal-2022-time}or generating hallucinatory content \cite{ji2023survey,shi2023replug,bang-etal-2023-multitask}. This impacts the dependability of LLMs and limits their broader practical use, as the alignment between LLM outputs and real-world information still requires further validation.
Augmenting LLMs with retrieval mechanisms to fetch relevant information from external sources has emerged as a promising approach to mitigate these issues \cite{khandelwal2019generalization, izacard2023atlas}.

Retrieval-augmented language models (LMs) typically operate using a retrieve-and-generate framework. This process begins by retrieving relevant documents based on the user's input. Subsequently, the model generates a comprehensive response that is conditioned on the information contained within these retrieved documents. This approach leverages the synergy between information retrieval and natural language generation, enhancing the model's ability to provide accurate and contextually relevant answers. \cite{chen-etal-2017-reading,guu2020realm,lewis2021retrievalaugmented,izacard-grave-2021-leveraging,sachan2021endtoend,lee2022need,jiang2022retrieval,izacard2023atlas,nakano2022webgpt,qian2023webbrain,lazaridou2022internetaugmented,shi2023replug}.

Standard RAG methods often involve a single retrieval step, which can be insufficient for complex problems requiring multi-step reasoning.  \cite{yoran2024making}. 
To address these limitations, various retrieval strategies such as Iterative Retrieval \cite{shao2023enhancing}, Recursive Retrieval \cite{trivedi2023interleaving,kim2023tree}, and Adaptive Retrieval \cite{jiang2023active,asai2023selfrag,yang2023autogpt,schick2023toolformer,zhang2023graphtoolformer} have been proposed. Among these, adaptive retrieval refines the RAG framework by enabling LLMs to actively determine the optimal moments and content for retrieval, thereby enhancing the efficiency and relevance of the sourced information. For example, Flare automates temporal retrieval by monitoring the confidence levels during the generation process, such as the probability of generated terms \cite{jiang2023active}. When this probability falls below a certain threshold, the retrieval system is activated to gather relevant information, thereby optimizing the retrieval cycle.

However, another significant challenge with naive RAG is its reliance on the user's original query for retrieval. Formulating precise and clear queries is difficult, leading to suboptimal retrieval effectiveness. Moreover, language complexity and ambiguity further complicate the process, as models may struggle with specialized vocabulary or ambiguous abbreviations.
To enhance retrieval effectiveness, query optimization strategies such as query expansion and query transformation have been developed. Query expansion enriches the content of the query by breaking down complex questions into simpler sub-queries or creating multiple parallel queries \cite{zhou2023leasttomost,dhuliawala2023chainofverification}. Query transformation involves rewriting or rephrasing the original query to improve retrieval effectiveness, using techniques like prompt engineering and hypothetical document generation \cite{ma2023query,peng2024large,gao2022precise,zheng2024step}.
These query optimization strategies are crucial for improving the effectiveness of RAG systems, ensuring they provide accurate and contextually appropriate responses.

While these existing methods are effective in many applications, they tend to focus on either query rewriting or retriever adaptation. Even when both aspects are considered, they are often addressed implicitly during the generation process. Moreover, in adaptive retrieval methods, the LM typically generates a response first and then decides whether additional retrieval is necessary based on the generated output. For instance, Flare automates temporal retrieval by evaluating the confidence in the generated terms \cite{jiang2023active}.

Given the robust semantic understanding capabilities of large language models (LLMs), we propose a hypothesis: can we assess the necessity of document retrieval before generating a response? This concept is inspired by the behavior of students during open-book exams. Faced with a question, students first understand the question, then evaluate their ability to answer it. If they can, they respond directly; if not, they consult their textbooks to gather the necessary information before crafting their final response. This two-step approach ensures that answers are both accurate and comprehensive. Applying this strategy to LLMs could potentially reduce the costs associated with calling APIs of black-box models, while maintaining or even enhancing response accuracy and relevance.

Building on this concept, this paper introduces \textit{Think-then-Act}, an accurate and efficient framework for retrieval augmentation, as illustrated in Figure~\ref{fig: frameword}. This framework incorporates a dual-phase evaluation and response process: (i) assessing the input query to determine if it is clear and complete and if it needs rewriting; (ii) evaluating the language model's capability to answer the query and whether additional information retrieval is necessary.

To validate the effectiveness of our proposed framework, we examine the performance of \textit{Think-then-Act} with gpt-3.5-turbo across five diverse datasets: HotPotQA \cite{yang-etal-2018-hotpotqa}, 2WikiMultihopQA \cite{ho-etal-2020-constructing}, StrategyQA \cite{geva2021did}, FEVER \cite{thorne-etal-2018-fever}, and a custom-built Chinese Poetry dataset. These datasets are chosen to comprehensively test various aspects of our approach, including multi-hop reasoning, commonsense reasoning, fact-checking, and domain-specific question answering. Our experimental results demonstrate that the \textit{Think-then-Act} framework significantly improves retrieval-augmented generation's performance, achieving higher accuracy and efficiency compared to existing baselines. Notably, the framework shows robust performance in both English and non-English contexts, highlighting its versatility and potential for broader applications.

\section{Related Work}

Our framework involves two modules of RAG: (i) query optimization within the context of RAG; and (ii) adaptive retrieval within the augmentation process of RAG.

\subsection{Query Optimization}

A key issue with Naive RAG is its dependence on the user's initial query \citep{gao2024retrievalaugmented}, often resulting in ineffective retrieval due to challenges in crafting clear questions and managing intricate or ambiguous language. 
Query transformation is an effective method for optimizing initial queries, which focuses on retrieving information using a modified query instead of the user's original query. 

Some studies use prompt engineering to enable LLM to generate a query based on the original one for subsequent retrieval \citep{jagerman2023query}. \citet{gao2022precise} generates hypothetical documents, which are presumed answers to the initial query. This approach emphasizes the similarity of embeddings between these generated answers rather than focusing on the similarity of embeddings related to the original problem or query. \citet{zheng2024step} using the Step-back Prompting method abstracts the initial query to formulate a broader, high-level conceptual question (step-back question). 
In addition to using LLM for rewriting, \citet{ma2023query} also specifically trained a smaller model to handle query rewriting tasks.

These methods enhance retrieval effectiveness; however, they assume that the input query always requires rewriting. Our approach introduces an evaluation step before rewriting, ensuring that the query is only modified if it is deemed incomplete or ambiguous.

\subsection{Adaptive Retrieval}

To improve factual accuracy, language models often rely on external knowledge via retrieval augmentation \citep{lewis2021retrievalaugmented}. Conventional retrieval-augmented generation (RAG) methods use a single retrieval step followed by generation, which can be insufficient for complex, multi-step reasoning tasks. Adaptive retrieval techniques optimize this process by allowing models to dynamically decide when and what to retrieve, enhancing both efficiency and relevance. 

One strategy is to add retrieval capabilities through the fine-tuning of a white-box generation model.
\citet{nakano2022webgpt} uses a reinforcement learning framework to train the GPT-3 model to autonomously use a search engine during text generation. It employs specific tokens to perform tasks such as making search queries, reviewing search results, and adding references, thus enhancing GPT-3's abilities with the help of external search engines.
\citet{asai2023selfrag} trained a flexible language model (LM) that can dynamically retrieve passages as required. This model uses special tokens, called 'reflection tokens,' classified into two types: 'retrieve' and 'critic,' to generate and review both the retrieved passages and its own outputs. By using these reflection tokens, the LM can be directed during the inference phase, allowing it to adapt its behavior to suit various task needs.
Additionally, some researchers use prompt engineering methods. Graph-Toolformer \citep{schick2023toolformer}, for instance, separates the retrieval process into specific stages, where LLMs actively use retrievers, utilize Self-Ask techniques, and apply few-shot prompts to start search queries.
Others \citep{jiang2023active} generates a preliminary answer first, then, based on whether the probability of the generated terms falls below a certain threshold, decides if additional information is needed before regenerating the response based on the initial result.

The generate-then-retrieve approach, while effective, is inefficient for queries that definitely need retrieval, as it introduces an extra generation step. We propose an approach where the model's capabilities are evaluated prior to generation, which achieves a balance between precision and efficiency in situations where absolute accuracy is not required.

\section{Methodology}

We present a dual-angle evaluated retrieval-augmented generation framework \textit{Think-then-Act}. This approach enhances both the query assessment and model capability evaluation processes. Figure~\ref{fig: frameword} provides an overview. This section first introduces the query assessment and rewriting process in Section~\ref{3.1}, followed by the model capability check and information retrieval in Section~\ref{3.2}.

\subsection{Initial Query Assessment}\label{3.1}

\subsubsection{Evaluation}

Accurate responses require clear and precise questions. Therefore, our first step involves evaluating the input query to determine whether it is clear and complete, incomplete, or ambiguous. Leveraging the inherent semantic understanding capabilities of large language models, we avoid the need for additional models for this evaluation. Instead, we use a prompting method that enables the model to self-assess the clarity and completeness of the input query, and the model categorizes the query as CLEAR AND COMPLETE, INCOMPLETE, or AMBIGUOUS.

\subsubsection{Rewriting}

If the evaluation categorizes the query as INCOMPLETE or AMBIGUOUS, the query requires rewriting. Utilizing the powerful generation capabilities of large language models, we employ a prompting method that enables the model to generate the revised queries itself.

\noindent
\textbf{INCOMPELTE} The model generates a more complete version of the query by filling in any missing information, ensuring it is clear and comprehensive.

\noindent
\textbf{AMBIGUOUS} The model resolves ambiguity by breaking down the query into multiple, straightforward sub-queries, each addressing a specific aspect of the original query.
Formally, the overall process of initial query $q_{init}$ assessment and rewriting, resulting in the model's final input $q_{final}$, can be summarized as follows:
\begin{equation}
  \label{eq:q_final}
  q_{\text{final}} =
  \left\{
  \begin{array}{lr}
    q_{\text{init}}  & \quad \text{if}\; \mbox{\small CLEAR AND COMPLETE} \\
    q_{\text{comp}}  & \quad \text{if}\; \mbox{\small INCOMPLETE} \\
    q_{\text{group}} = & \{q^1, q^2, \ldots, q^k\} \quad \text{if} \; \mbox{\small AMBIGUOUS}
  \end{array}
  \right.
\end{equation}

\subsection{Model Capability Check}\label{3.2}

After completing the initial query assessment and obtaining the final input $q_{final}$, the next step involves evaluating the LM's capability to answer $q_{final}$. We propose two methods for this evaluation:

\noindent
\textbf{Direct Decision:} In this straightforward approach, the LM directly outputs either RETRIEVAL or NO RETRIEVAL. This binary decision indicates whether the LM needs additional information to answer the query effectively.

\noindent
\textbf{Confidence Score:} This method involves the LM generating a confidence score, denoted as $\beta$, which represents its confidence level in answering the question. By comparing this score to a predefined threshold $\beta'$, we can dynamically decide whether retrieval is necessary. $\beta < \beta'$, indicating that the model lacks sufficient confidence to answer the query on its own, so retrieval is required. $\beta >= \beta'$, suggesting that the model is confident in its ability to provide an accurate response without additional information, so retrieval is not needed. $\beta' \in [0,1]$. When $\beta' = 0$ it means that retrieval is never performed. When $\beta' = 1$ it means that retrieval is performed for every $q_{final}$.

\subsubsection{Information Retrieval}

If the model determines that additional information is necessary, we proceed with the information retrieval step. Search engines possess features that large language models (LLMs) lack, such as the ability to be easily and quickly updated \citep{kasai2024realtime}. We use the Google search engine and Wikipedia-API(wiki) as the retriever to obtain relevant documents $D$, 2 example in Table~\ref{tab:prompts}. Detailed settings for the retrieval process are described in Section~\ref{4.2}.
\begin{equation}
    \label{eq:y_out}
    y_{\text{output}} = 
    \left\{
    \begin{array}{ll}
       LM(\: q_{\text{final}} \: ) & \text{if } \:  \beta \geq \beta' \\
       LM(\: [\: D, \: q_{\text{final}} \: ] \: ) & \text{if } \:  \beta < \beta'
    \end{array}
    \right.
\end{equation}

\begin{table*}[t]
    \resizebox{\linewidth}{!}{
    \begin{tabular}{l}
         \hline
            \textbf{Question: Is popular science used to peer review papers?} \\
            The question is: ambiguous. \\
            Sub-Question: What is popular science? \\
            Probability of correct answer is: 0.5. \\
            Sub-Answer: Popular science is a simplified version of scientific work. \\
            Sub-Question: What types of documents does peer review use to verify papers? \\
            Probability of correct answer is: 0.6. \\
            Sub-Answer: Peer review uses detailed scientific information to verify papers. \\
            So the final answer is: False. \\
            \hline
            \textbf{Question: Does a lapidary work with items that are studied by geologists?} \\
            The question is: ambiguous. \\
            Sub-Question: What are the materials a lapidary works with? \\
            Probability of correct answer is: \colorbox{yellow}{0.4.} \\
            Sub-Answer: \colorbox{green}{Quartz is a popular material for lapidary because it is relatively easy to work with and comes in various colours and patterns. As it is .... }\\
            Sub-Question: What do geologists study? \\
            Probability of correct answer is: 0.5. \\
            Sub-Answer: Lapidarists work with stone, minerals and gemstones. \\
            So the final answer is: True. \\
            \hline
        \hline
    \end{tabular}
    }
    \caption{Think-then-Act on StrategyQA: 2 samples}
    \label{tab:prompts}
\end{table*}

~
\section{Experimental Setup}

\subsection{Task Settings}

To thoroughly evaluate the capabilities of the \textit{Think-then-Act} framework, we selected tasks and datasets designed to comprehensively test various aspects of our approach. Specifically, we chose three sub-tasks: Multihop QA, Commonsense Reasoning, Fact Checking and Domain QA.

\subsubsection{MultihopQA}

Two multihop QA datasets are used for evaluation. (i) HotPotQA \citealp{yang-etal-2018-hotpotqa}: consists of complex questions that require multi-hop reasoning, where the answer to a question requires synthesizing information from multiple documents. For our evaluation, we use the full test set, ensuring a comprehensive assessment of our framework capabilitues of multihop reasoning. (ii) 2WikiMultihopQA \citealp{ho-etal-2020-constructing} is a multi-hop question-answering dataset that exploits the structured format in Wikidata and uses logical rules to create questions. By evaluating on this dataset, we aim to test the proficiency of our framework in handling structured information and executing logical inference.

\subsubsection{Commonsense Reasoning}

Commonsense reasoning requires a blend of world knowledge and commonsense understanding to generate accurate answers. For this purpose, we utilize the StrategyQA dataset \cite{geva2021did}, which consists of crowd sourced yes/no questions, such as “Would a pear sink in water?”. The final answers are extracted and matched against the gold standard answers using exact match to evaluate the performance of our framework in commonsense reasoning tasks.

\subsubsection{Fact Checking}

We also employ the FEVER dataset \cite{thorne-etal-2018-fever} for fact verification tasks. This dataset categorizes claims as "SUPPORTS", "REFUTES", or "NOT ENOUGH INFO" based on evidence paragraphs extracted from Wikipedia. To ensure a challenging evaluation, we sample a balanced set of instances where GPT-3's chain-of-thought (CoT) method makes both correct and incorrect predictions. This approach allows us to rigorously test the model's ability to verify facts using evidence-based reasoning.

\subsubsection{Domain QA}

For the domain-specific question answering task, we utilize a custom-built dataset focused on \textbf{Chinese Poetry}. This dataset was developed in response to recurring issues with existing QA models, such as ChatGPT, which often incorrectly match poetry verses with their titles and authors. Our analysis revealed two primary reasons for these errors: firstly, the models may possess accurate parametric knowledge but still generate incorrect answers; secondly, they may lack the requisite information for obscure poetry verses. Consequently, this dataset is ideally suited to evaluate the effectiveness of our framework. Additionally, this allows us to test the effectiveness of our framework in the Chinese language context, extending its applicability beyond just English.

Our custom dataset comprises 9,791 poetry verses from 60 different poets, providing a comprehensive basis for testing. This dataset enables us to thoroughly assess the ability of our framework to handle both common and obscure queries in the domain of classical Chinese poetry.

\begin{figure*}[ht]
  \centering
  \includegraphics[width=1.00\linewidth]{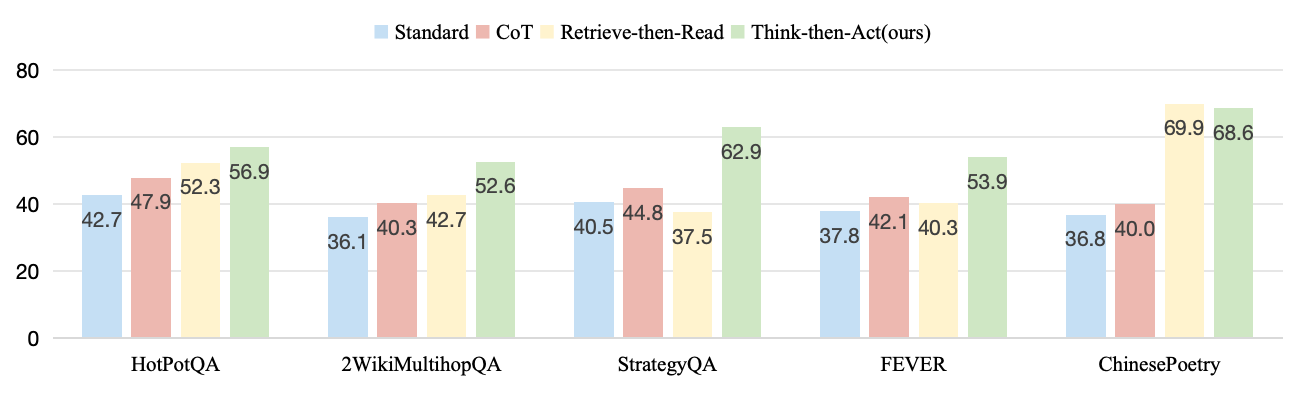}
  \caption{Overall results using the EM metric.Think-then-Act parameter $\beta'=0.5$.}
  \label{fig: overall_results}
\end{figure*}

\subsection{Retriever Details}\label{4.2}

In the information retrieval step, we use two systems to obtain relevant information:

\textbf{Wikipedia-API}(wiki): For the final query $q_{final}$, we search through Wikipedia and select the top sentences from the relevant Wikipedia pages. This approach leverages the structured and comprehensive nature of Wikipedia to provide accurate and detailed information.

\textbf{Google Search Engine}: For queries that can be directly answered, such as "Where is the capital of China?", Google searches often present direct "$answer \; boxes$". We utilize these explicit answers for straightforward questions. For more complex queries, Google provides "$organic \; results$" as the main search output.

For wiki and the second case of Google, we select the top 3 most similar to the query ranked by the pre-trained Sentence BERT model\cite{reimers2019sentencebert} as context.

\subsection{Baseline}\label{4.3}

To provide a comprehensive evaluation of our framework, we compare it against the following baselines. (i) \textbf{Standard Prediction (Standard)}: This baseline involves directly predicting the label based on the input, utilizing the same number of in-context learning examples as our framework. (ii) \textbf{Original Chain-of-Thought (CoT)}\cite{wei2023chainofthought}: This approach predicts the label after generating an explanatory chain-of-thought. It helps in understanding the model's reasoning process and its impact on the final prediction. (iii) \textbf{Retrieve-then-Read}: This is the standard retrieval-augmented method where retrieved documents are concatenated with the question to form the input. This baseline allows us to measure the performance gains from our dual-focus approach compared to traditional retrieval-augmented methods.

\begin{table*}[t]
    \begin{tabular}{l cc|cc|cc|cc|cc}
    \toprule
        \textbf{Datasets}  & \multicolumn{2}{c|}{\textbf{HotPotQA}}  & \multicolumn{2}{c|}{\textbf{2Wiki.}}  & \multicolumn{2}{c|}{\textbf{StrategyQA}} & \multicolumn{2}{c|}{\textbf{FEVER}}  & \multicolumn{2}{c}{\textbf{ChinesePoetry}} \\
        \textbf{Metrics}            & \textbf{EM} & \textbf{F1} & \textbf{EM} & \textbf{F1} & \textbf{EM} & \textbf{F1} & \textbf{EM} & \textbf{F1} & \textbf{EM} & \textbf{F1} \\
        \hline
        Standard & 42.7 & 51.3 & 36.1 & 54.5 & 40.5 & 57.2 & 37.8 & 50.4 & 36.8 & 39.3 \\
        CoT & 47.9 & 59.7 & 40.3 & 60.2 & 44.8 & 60.1 & 42.1 & 57.4 & 40.0 & 49.2 \\
        Retrieve-then-Read & 52.3 & \textbf{66.4} & 42.7 & 69.3 & 37.5 & 50.7 & 40.3 & \textbf{59.2} & \textbf{69.9} & \textbf{76.2} \\
        \text{Think-then-Act(ours)} & \textbf{56.9} & 65.8 & \textbf{52.6} & \textbf{69.7} & \textbf{62.9} & \textbf{71.2} & \textbf{53.9} & 55.7 & 68.6 & 70.1 \\
        \bottomrule
    \end{tabular}
    \caption{\label{table:results} Comparison between Think-then-Act ($\beta'=0.5$) and baselines on all datastes using EM \& F1. }
\end{table*}

\section{Experimental Results}

We first report the overall results across the selected tasks and datasets, comparing the performance of the \textit{Think-then-Act} framework with all the baselines introduced in Section~\ref{4.3}. We then conduct ablation experiments to study the efficacy of various design choices within our method. This structured analysis allows us to thoroughly evaluate the strengths and areas for improvement in our approach.

\subsection{Comparison with Baselines}

Figure~\ref{fig: overall_results} displays the performance comparison of the \textit{Think-then-Act} framework against the baselines across various tasks and datasets. Our framework generally outperforms the baselines, indicating its superior capability in enhancing retrieval-augmented generation. 

\begin{figure}[t]
  \centering
  \includegraphics[width=1.00\linewidth]{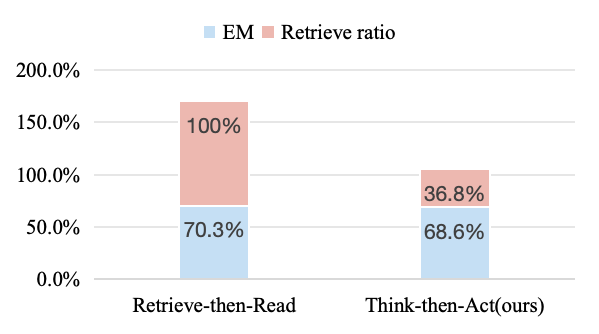}
  \caption {Comparison of Think-then-Act and Retrieval-then-Read($\beta'=0.5$) on the Chinese Poetry dataset: generation accuracy(blue) and retrieval ratio(red).}
  \label{fig: directR_ous}
\end{figure}

\noindent
\textbf{Comparisons on StrategyQA:} The most notable improvement is observed in StrategyQA, shown in Table~\ref{table:strategyQA_FEVER}. The Chain of Thought(CoT)method \citep{wei2023chainofthought}, which involves deeper question analysis, outperforms direct retrieval methods. This is directly related to the characteristics of the StrategyQA dataset. For commonsense reasoning tasks, deeply analyzing and understanding the question is more crucial than acquiring additional information, few example in Table~\ref{tab:prompts}. This trend is similarly observed in the FEVER dataset, where accurate fact verification benefits more from a thorough understanding of the query rather than from additional data.

\begin{table}[t]
    \resizebox{\linewidth}{!}{
    \begin{tabular}{l cc|cc}
        \hline
        \textbf{Datasets}  & \multicolumn{2}{c|}{\textbf{StrategyQA}} & \multicolumn{2}{c}{\textbf{FEVER}}   \\
        \textbf{Metrics}   & \textbf{EM} & \textbf{F1}                & \textbf{EM} & \textbf{F1} \\
        \hline
        Retrieve-then-Read & 37.5 & 50.7 & 40.3 & \textbf{59.2} \\
        \text{Think-then-Act(ours)} & \textbf{62.9} & \textbf{71.2} & \textbf{53.9} & 55.7 \\
        \hline
    \end{tabular}}
    \caption{\label{table:strategyQA_FEVER}Comparison between Think-then-Act ($\beta'=0.5$) and Retrieve-then-Read on all StrategyQA and FEVER}
\end{table}

\noindent
\textbf{Comparisons on ChinesePoetry:} On our custom ChinesePoetry dataset, our method performs comparably to the Retrieve-then-Read baseline. This can be attributed to the clarity and completeness of the questions in this dataset, where additional information retrieval significantly enhances accuracy. However, unlike the baseline method that retrieves information for all queries, our approach first assesses the model's capability before deciding whether retrieval is necessary. As shown in Figure~\ref{fig: directR_ous}, our method retrieves information for only 36.8\% of the questions, achieving the same effectiveness as retrieving for 100\% of the queries. This selective retrieval significantly reduces computational costs.

We report all metrics for the every baselines in Table~\ref{table:results}, highlight the performance metrics (EM and F1 scores) for different methods across various datasets. Our \textit{Think-then-Act} framework consistently demonstrates superior performance, particularly in tasks requiring complex reasoning and fact verification. Notably, it achieves the highest EM scores in HotPotQA (56.9), 2WikiMultihopQA (52.6), StrategyQA (62.9), and FEVER (53.9), showcasing its robustness and adaptability. The framework's comparable performance in the ChinesePoetry dataset (EM: 68.6) against the Retrieve-then-Read baseline (EM: 69.9) further illustrates its efficiency in handling domain-specific tasks with reduced computational overhead.

\begin{figure}[t]
  \centering
  \includegraphics[width=1.00\linewidth]{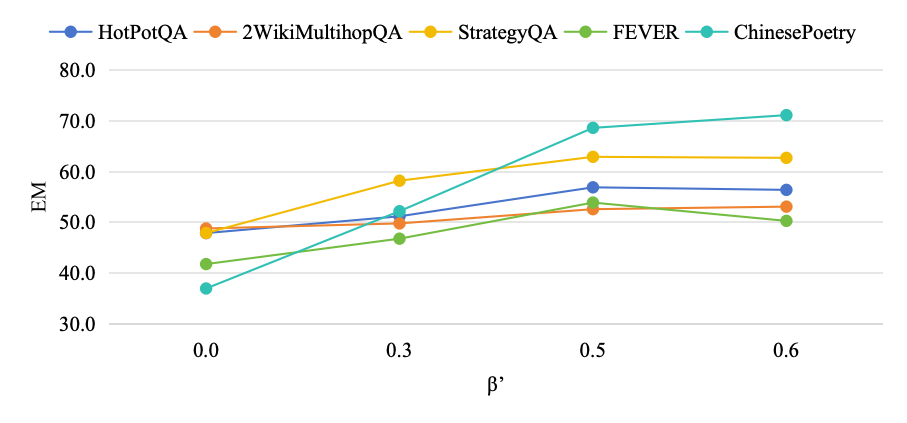}
  \caption {the Exact Match (EM) scores across various datasets with different $\beta'$ values}
  \label{fig: beta}
\end{figure}

\begin{figure}[t]
  \centering
  \includegraphics[width=1.00\linewidth]{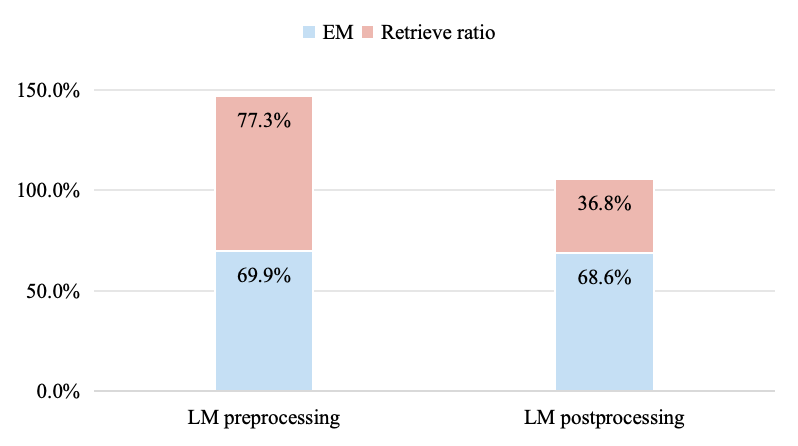}
  \caption {Comparison of Think-then-Act($\beta'=0.5$, LM postprocessing) and FLARE(LM preprocessing) on the Chinese Poetry dataset: generation accuracy(blue) and retrieval ratio(red).}
  \label{fig: pre_post}
\end{figure}

\subsection{Ablation Study}

Our framework \textit{Think-then-Act}, primarily consists of two components: query assessment and model capability check. Unlike traditional approaches, we do not omit these parts separately to demonstrate their effectiveness, as previous studies have already established their importance. Instead, our ablation study focuses on two main experiments to validate the design choices and their impact on performance.

\noindent
\textbf{Impact of Different $\boldsymbol{\beta'}$:} As mentioned in Section~\ref{3.2}, we examine how varying the threshold $\beta'$ for model confidence affects the results. This helps us understand the optimal threshold for balancing retrieval necessity and model confidence.

Figure~\ref{fig: beta} displays the Exact Match (EM) scores across various datasets with different $\beta'$ values. As observed, the performance improves significantly when $\beta'$ is increased from $0.0$ to $0.5$. Beyond $0.5$, the improvement plateaus, indicating diminishing returns. Therefore, we set $\beta'$ to $0.5$ for optimal performance, balancing the trade-off between retrieval and self-reliance of the model.

\noindent
\textbf{Preprocess and Postprocess:} We compare our framework with a method inspired by FLARE \citep{jiang2023active}, where the LM generates a temporary next sentence and checks the token probabilities before deciding on retrieval. We refer to this as the \textit{LM preprocess} approach. In contrast, our framework first assesses whether retrieval is needed and then generates the response, which we term as the \textit{LM postprocess} approach.

Figure~\ref{fig: pre_post} illustrates the comparison between LM preprocessing and LM postprocessing. The results show that our postprocessing approach achieves a comparable EM score (68.6\%) to the preprocessing approach (69.9\%), but with a significantly lower retrieval ratio (36.8\% vs. 77.3\%). This indicates that our method is more efficient, reducing the number of retrievals required while maintaining similar performance. Consequently, this leads to faster processing times and enhanced resource efficiency.

\section{Conclusion}

This paper presents the \textit{Think-then-Act} framework, enhancing retrieval-augmented generation by combining query transformation and model capability assessment. Our approach evaluates query clarity and model confidence, triggering retrieval only when necessary, improving accuracy, and optimizing resources. Experiments on datasets including MultihopQA, Commonsense Reasoning, FEVER, and a custom Chinese poetry dataset show significant improvements over baselines. The framework proves effective in both English and non-English contexts. Ablation studies confirmed the optimal model confidence threshold and highlighted efficiency gains from our approach. The \textit{Think-then-Act} framework offers a robust solution for enhancing retrieval-augmented generation, paving the way for more accurate and efficient LLM applications. Future work will refine query assessment and extend the framework to additional languages and domains.

\section{Limitations}

While the \textit{Think-then-Act} framework has demonstrated promising results, several limitations need to be addressed in future research.
Firstly, our study exclusively utilized black-box models, such as GPT-3.5, which necessitate API calls for each interaction. This approach incurs significant costs and poses potential security risks due to the transmission of data over external servers. Inspired by \citep{asai2023selfrag}, future work could focus on fine-tuning a white-box model based on our framework. This would enable local deployment, thereby reducing costs and enhancing data security by processing all information internally.
Secondly, although we conducted experiments on five distinct datasets, each experiment was isolated to a single type of dataset. This approach does not fully capture the versatility and robustness of our framework across mixed-type scenarios. Future research should investigate the performance of our framework on more diverse and comprehensive datasets that incorporate various types of questions and contexts within a single dataset. This would provide a more rigorous validation of the framework's effectiveness and adaptability in real-world applications.

\bibliography{latex/think_then_act}

\appendix

\newpage
\section{Appendix}
\begin{table*}[t]
    \resizebox{\linewidth}{!}{
    \begin{tabular}{l}
         \hline
            \textbf{Question: Could Carl Friedrich Gauss speak to someone 100 miles away?} \\
            The question is: ambiguous. \\
            Sub-Question: What device allows people to speak to each other even if they are 100 miles apart? \\
            Probability of correct answer is: 0.8. \\
            Sub-Answer: Carl Friedrich Gauss was born in 1777. \\
            Sub-Question: When was \#1 invented? \\
            Probability of correct answer is: 0.8. \\
            Sub-Answer: Speaking to someone 100 miles away requires a telephone. \\
            Sub-Question: When did Carl Friedrich Gauss die? \\
            Probability of correct answer is: 0.8. \\
            Sub-Answer: The telephone was invented in 1876. \\
            So the final answer is: False. \\
            \hline
            \textbf{Question: Did Jackson 5 members exceed number in The Osmonds?} \\
            The question is: ambiguous. \\
            Sub-Question: How many members did The Jackson 5 have? \\
            Probability of correct answer is: \colorbox{yellow}{0.3}. \\
            Sub-Answer: The group was founded in 1964 in Gary, Indiana, and originally consisted of brothers Jackie, Tito, Jermaine, Marlon and Michael. They were managed by their .... \\
            Sub-Question: How many members did The Osmonds have? \\
            Probability of correct answer is: \colorbox{yellow}{0.3}. \\
            Sub-Answer: The Osmonds then and now · Donny Osmond · Marie Osmond · Jay Osmond · Merrill Osmond. Get info without leaving the page. · Wayne Osmond · Alan Osmond. \\
            So the final answer is: False. \\
            \hline
            \textbf{Question: Is popular science used to peer review papers?} \\
            The question is: ambiguous. \\
            Sub-Question: What is popular science? \\
            Probability of correct answer is: 0.5. \\
            Sub-Answer: Popular science is a simplified version of scientific work. \\
            Sub-Question: What types of documents does peer review use to verify papers? \\
            Probability of correct answer is: 0.6. \\
            Sub-Answer: Peer review uses detailed scientific information to verify papers. \\
            So the final answer is: False. \\
            \hline
            \textbf{Question: Does a lapidary work with items that are studied by geologists?} \\
            The question is: ambiguous. \\
            Sub-Question: What are the materials a lapidary works with? \\
            Probability of correct answer is: \colorbox{yellow}{0.4.} \\
            Sub-Answer: \colorbox{green}{Quartz is a popular material for lapidary because it is relatively easy to work with and comes in various colours and patterns. As it is .... }\\
            Sub-Question: What do geologists study? \\
            Probability of correct answer is: 0.5. \\
            Sub-Answer: Lapidarists work with stone, minerals and gemstones. \\
            So the final answer is: True. \\
            \hline
            \textbf{Question: Could Carl Friedrich Gauss speak to someone 100 miles away?} \\
            The question is: ambiguous. \\
            Sub-Question: What device allows people to speak to each other even if they are 100 miles apart? \\
            Probability of correct answer is: \colorbox{yellow}{0.1.} \\
            Sub-Answer: Start with a handheld. I like Yaesus. \\
            Sub-Question: When was \#1 invented? \\
            Probability of correct answer is: \colorbox{yellow}{0.4.} \\
            Sub-Answer: 1 (one, unit, unity) is a number representing a single or the only entity. 1 is also a numerical digit and represents a single unit of counting or .... \\
            Sub-Question: When did Carl Friedrich Gauss die? \\
            Probability of correct answer is: \colorbox{yellow}{0.3.} \\
            Sub-Answer: Carl Friedrich Gauss (born April 30, 1777, Brunswick [Germany]—died February 23, 1855, · Gauss was the only child of poor parents. · Gauss's first .... \\
            So the final answer is: False. \\
            \hline
            \textbf{Question: Can you listen to the entire iTunes song catalog in one year?} \\
            The question is: ambiguous. \\
            Sub-Question: How many songs are on iTunes? \\
            Probability of correct answer is: 0.8. \\
            Sub-Answer: iTunes has around 43 million songs as of 2017. \\
            Sub-Question: What is the average song length? \\
            Probability of correct answer is: 0.9. \\
            Sub-Answer: The average length of a song is 3 minutes. \\
            Sub-Question: What is \#1 multiplies by \#2? \\
            Probability of correct answer is: 0.6. \\
            Sub-Answer: There are 525,600 minutes in a year. \\
            So the final answer is: False. \\
            \hline
            \textbf{Question: Can you listen to the entire iTunes song catalog in one year?} \\
            The question is: ambiguous. \\
            Sub-Question: How many songs are on iTunes? \\
            Probability of correct answer is: \colorbox{yellow}{0.0.} \\
            Sub-Answer: If you go to the "Songs" window and reenable the status bar by selecting "View" then "Show Status Bar" you will get the total count at the .... \\
            Sub-Question: What is the average song length? \\
            Probability of correct answer is: \colorbox{yellow}{0.2.} \\
            Sub-Answer: Nowadays, songs average around 3:15/3:30 which shows a decrease in length by up to 60 seconds. On top of that, we're constantly seeing way more .... \\
            Sub-Question: What is \#1 multiplies by \#2? \\
            Probability of correct answer is: 0.7. \\
            Sub-Answer: 1/2. \\
            So the final answer is: False. \\
            \hline
        \hline
    \end{tabular}
    }
    \caption{Think-then-Act on StrategyQA: 10 samples}
    \label{tab:my_label}
\end{table*}

\end{document}